# LogSyn: A Few-Shot LLM Framework for Structured Insight Extraction from Unstructured General Aviation Maintenance Logs

Devansh Agarwal[1*], Maitreyi Chatterjee[1*], Biplab Chatterjee[2]
[1]Cornell University, Ithaca, NY
[2]Ground Handling, AI Airport Services Ltd, Kolkata, India
* Equal Contribution

## ABSTRACT

Aircraft maintenance logs hold valuable safety data but remain underused due to their unstructured text format. This paper introduces LogSyn, a framework that uses Large Language Models (LLMs) to convert these logs into structured, machine-readable data. Using few-shot in-context learning on 6,169 records, LogSyn performs Controlled Abstraction Generation (CAG) to summarize problem–resolution narratives and classify events within a detailed hierarchical ontology. The framework identifies key failure patterns, offering a scalable method for semantic structuring and actionable insight extraction from maintenance logs. This work provides a practical path to improve maintenance workflows and predictive analytics in aviation and related industries.

## NOMENCLATURE

**LLM:** Large Language Model
**GA:** General Aviation
**JSON:** JavaScript Object Notation
**CAG:** Controlled Abstraction Generation

## 1. INTRODUCTION

In aviation, safety and reliability are paramount. This safety culture relies on meticulous record-keeping, with maintenance logs serving as a rich source of high-dimensional data for failure prevention, airworthiness, and fleet optimization. These logs capture decades of operational experience through detailed problem and repair records. However, their unstructured, jargon-heavy text locks away critical insights, limiting large-scale analysis and predictive modeling. As a result, operators remain constrained to reactive maintenance, higher costs, and lost opportunities for proactive safety gains.

Advancements in machine learning and deep learning [1, 2, 3, 4, 11] have introduced powerful new paradigms for data analysis. Traditional Natural Language Processing (NLP) offers initial pathways to extract keywords but often fail to perform robust causal-symptom mapping or capture the semantic nuances embedded in the narratives. For instance, knowing that "gasket" and "leak" are frequent terms is less valuable than understanding that "a leaking rocker cover gasket was the cause of an in-flight engine issue." Modern, transformer-based LLMs like GPT-4 and Gemini mark a paradigm shift [9, 10], offering the ability to perform deep semantic understanding and structured data extraction.

This paper introduces LogSyn, a novel framework that directly addresses this challenge by leveraging LLMs to systematically translate unstructured maintenance narratives into a schematized, machine-readable format. This work utilizes the public Aircraft Historical Maintenance Dataset (2012–2017) [5], a dataset of GA logs with both "Problem" and "Action Taken" text fields, totaling 6,169 records. LogSyn's primary contribution is a cost-effective and scalable methodology that transforms these narratives into concise summaries and assigns them to a granular, data-driven fault ontology, thereby unlocking the latent value within these critical records.

## 2. METHODOLOGY

### 2.1 Dataset

We use Aircraft Historical Maintenance Dataset (2012–2017) from Kaggle, which contains 6,169 maintenance entries from general aviation aircraft (primarily Cessna 172s). Each record has two free-text fields: Problem (symptom description) and Action Taken (corrective steps). We have expanded the dataset by getting these records' ontology classified by Gemini and then later on hand-correcting the labels which we thought are not correct. These labels are used in Section 3.5.

### 2.2 LogSyn System Architecture

The LogSyn pipeline comprises five stages:

a) **Preprocessing:** Clean whitespace, normalize domain-specific abbreviations, and concatenate the problem-action text.
b) **Prompt Construction:** Construct a meta-learning context via a few-shot prompt with task instructions and 2–3 exemplars for in-context conditioning.

c) **LLM Inference:** Query an LLM (GPT-4 or Gemini) with deterministic settings (temp. = 0.1) to generate a schematized representation.
d) **Postprocessing:** Parse and validate the JSON output, flagging structural anomalies for review.
e) **Aggregation:** Collect structured entries for macro-level analysis and visualization.

### 2.3 Few-Shot Prompt Engineering

We craft prompts that guide the LLM to produce a JSON object serving as a schematized representation of the event, with fields: summary problem, summary action, failed component, and category. The category field uses a two-level, domain-specific ontology derived from the maintenance logs to provide more specific insights. **Example Prompt Snippet:** Problem: #2 & 4 CYL ROCKER COVER GASKETS ARE LEAKING. Action Taken: REMOVED & REPLACED GASKETS.

**Output:**

{ "summary_problem": "Rocker cover gasket leaks in cylinders 2 and 4 were reported.", "summary_action": "The leaking rocker cover gaskets were replaced on cylinders 2 and 4.", "failed_component": "Rocker Cover Gaskets (Cyl 2 & 4)", "category": "Powerplant - Sealing & Gaskets" }

This exemplar-based conditioning approach [6] avoids costly fine-tuning while ensuring consistency and granularity in the output format.

## 3. RESULTS AND DISCUSSION

### 3.1 Qualitative Analysis

We manually reviewed a sample of model outputs, confirming accurate Controlled Abstraction Generation (CAG) and classification against our domain-specific ontology. Table 1 goes over sample model outputs:

**Table 1: Sample Model Outputs with Enhanced Categorization**

| **Problem** | **Action taken** | **Category** |
|---|---|---|
| Had engine choke & briefly lose power on departure. | Performed engine run-up, found cyl 2 lower plug fouled. | Ignition System - Component Failure |
| #4 rocker cover is leaking. | Removed & replaced #4 rocker cover gasket. | Powerplant - Sealing & Gaskets |
| Had engine choke & briefly lose power on departure. | Stop drilled crack. | Powerplant - Structural Components |

### 3.2 Quantitative Analysis: A Granular Maintenance Ontology

A key outcome of applying LogSyn is the ability to develop a granular, data-driven maintenance ontology. Unlike predefined taxonomies, the categories in Table 2 were derived directly from recurring themes identified by the LLM within the dataset. This bottom-up approach ensures that the ontology is highly relevant to the specific operational context of GA aircraft, capturing the most frequent and critical maintenance events [7]. The rationale is to create a classification scheme that reflects the reality of day-to-day maintenance rather than imposing a generic structure. This process revealed specific, high-frequency patterns, particularly within powerplant maintenance. Figure 1 and Table 2 shows the distribution of the logs in the 8 maintenance categories.

**Table 2: Data-Driven Maintenance Ontology Derived from Logs**

| **Category (Distribution Quantity)** | **Description** | **Example Faults Logs** |
|---|---|---|
| Powerplant - Mechanical (553) | Issues related to the core mechanical components of the engine. | Low compression, Piston/ring failure, Sticking valves |
| Powerplant - Sealing & Gaskets (3454) | Failures of seals, gaskets, and O-rings leading to fluid leaks. | Leaking rocker cover gasket,Intake manifold leak, Oil seal failure |
| Powerplant - Structural Components (846) | Cracks, wear, or failure of non-moving engine parts. | Cracked engine baffle, Worn engine mount, Broken bracket |
| Powerplant - Fasteners & Hardware (588) | Issues involving loose, missing, or broken screws, clamps, and rivets. | Loose rocker cover screws, Broken hose clamp, Sheared rivets |
| Ignition System - Component Failure (76) | Malfunctions within the ignition circuit | Fouled spark plug, Magneto failure, Faulty ignition lead |
| Fuel System - Delivery & Control (50) | Problems with fuel flow, mixture, and idle settings. | Fuel servo malfunction, Clogged injector nozzle, Incorrect idle mixture |
| Performance - Operational Issue (403) | Discrepancies reported by pilots without a specific component identified | Rough running engine, Power loss, Hard start, Vibration |
| Servicing - General Maintenance (199) | Routine inspections, cleaning, and non-failure related tasks. | FOD removal, Engine wash, Scheduled compression check |

### 3.3 Visual Analysis

The true power of this structured data is revealed through visualization. By converting unstructured text into discrete categories, we can analyze the relationships between problems and actions at scale. A Sankey diagram (Fig. 2) was generated to illustrate these problem-to-action pathways. The width of each flow is proportional to its frequency, providing an immediate, quantitative view of the most critical maintenance trends. For instance, the diagram makes it unequivocally clear that issues within 'Powerplant - Sealing & Gaskets' are the single largest driver of 'Component Replacement' actions. This level of insight is fundamental for prioritizing maintenance training, inventory management, and predictive modeling, yet it remains completely obscured in raw text logs.

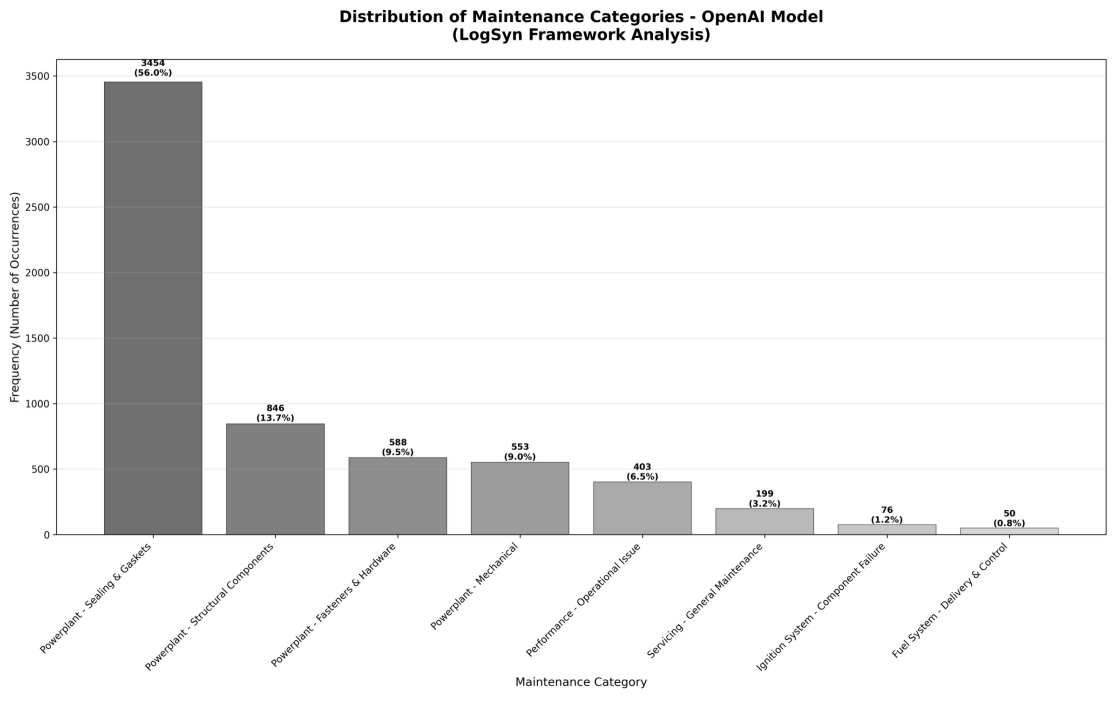

**Fig. 1: Distribution of Maintenance Categories**

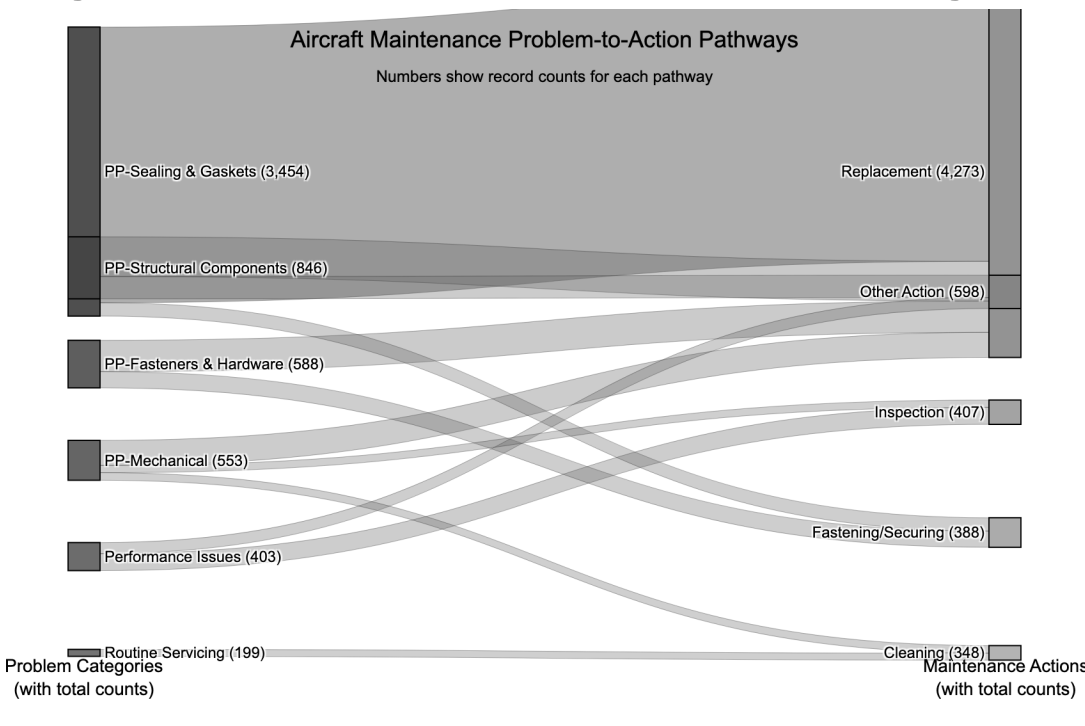

**Fig. 2: Aircraft Maintenance: Problem-to-Action Pathways**

### 3.4 LLM-as-a-Judge-Evaluation

To provide a scalable evaluation without manual annotation [8], we adopted an LLM-as-a-Judge approach, a programmatic, model-based heuristic for assessing semantic fidelity. A second LLM rated outputs on a 1–5 Likert scale for summary accuracy, component correctness, and category relevance. As shown in Table 3, this evaluation yielded high average scores across all criteria, with a mean of 4.7 for summary accuracy and 4.8 for category relevance. This suggests that the generated outputs are structurally and semantically accurate and contextually appropriate.

**Table 3: LLM-as-a-Judge Results**

| **Metric** | **Score** |
|---|---|
| Summary Accuracy | 4.7 |
| Component Accuracy | 4.5 |
| Category Relevance | 4.8 |

### 3.5 Comparison with Baseline

LogSyn's classification performance is evaluated against three baselines: (1) a zero-shot LLM classifier, (2) a rule-based NER (regex + spaCy), and (3) a supervised NER (fine-tuned BERT with cosine similarity). Table 4 summarizes the results. LogSyn's few-shot approach outperformed all baselines across macro-averaged metrics. While overall accuracy improved modestly over the zero-shot model (0.9021 vs. 0.8899), macro-precision and macro-F1 rose by 10% and 7%, respectively. Macro-averaging, which treats all classes equally, highlights LogSyn's strength in handling rare events. These results show that few-shot, in-context learning delivers more consistent and accurate classification than zero-shot or rule-based methods.

**Table 4: LogSyn Baseline Comparisons**

| **Metric** | **Baseline (Zero-Shot)** | **LogSyn (Few-Shot)** | **Rule-Based NER** | **BERT** |
|---|---|---|---|---|
| Accuracy | 0.8899 | **0.9021** | 0.7992 | 0.7314 |
| Precision (Macro) | 0.6427 | **0.7455** | 0.6326 | 0.53228 |
| Recall (Macro) | 0.7428 | **0.7779** | 0.5067 | 0.4465 |
| F1-Score (Macro) | 0.6891 | **0.7614** | 0.4894 | 0.4309 |

### 3.6 Limitations

LogSyn's performance is influenced by prompt design and example selection.

- Prompt Sensitivity: Accuracy varies by 2-4% across prompts, requiring domain

expertise for optimal phrasing, unlike supervised learning methods.

- Example Bias: Few-shot examples affect rare class performance. Bad examples can lead to possible overfitting to specific writing styles.

## 4. CONCLUSION

LogSyn provides a reproducible framework for the semantic structurization of free-text GA maintenance logs using off-the-shelf LLMs. Through Controlled Abstraction Generation and the application of a data-derived, hierarchical ontology, it unlocks granular, actionable insights. Crucially, the transformation from unstructured text to a structured format enables the quantitative analysis needed to identify and rank systemic maintenance trends. This allows stakeholders to focus resources on the most prevalent and critical issues, such as powerplant gasket failures, thereby paving the way for integration into predictive maintenance dashboards and enabling longitudinal analysis of fault taxonomies. In practice, LogSyn can be deployed as an API or integrated module within existing maintenance systems, automatically processing incoming log entries to provide real-time fault summaries and trend analytics.

## 5. FUTURE WORK

Some future work possibilities include:

- **Temporal and Root Cause Analysis:** The structured data enables longitudinal analysis to identify temporal trends, such as seasonal failure patterns or the emergence of recurring issues across a fleet.
- **Interactive Systems:** Developing an interactive real-time dashboard or digital-twins for maintenance technicians can significantly reduce the errors and maintenance time.
- **Advanced LLM Methodologies:** While few-shot prompting is cost-effective, future research should explore fine-tuning models using direct preference optimization which require lesser-compute and are faster than traditional reinforcement learning with human feedback.
- **Cross-Industry Adaptation:** The LogSyn framework could be adapted for other high-safety industries that rely on unstructured text logs, such as railways, maritime, or the energy sector, to improve their maintenance and safety.